# A Review on the Application of Natural Computing in Environmental Informatics

Andreas Kamilaris[1]

**Abstract:** Natural computing offers new opportunities to understand, model and analyze the complexity of the physical and human-created environment. This paper examines the application of natural computing in environmental informatics, by investigating related work in this research field. Various nature-inspired techniques are presented, which have been employed to solve different relevant problems. Advantages and disadvantages of these techniques are discussed, together with analysis of how natural computing is generally used in environmental research.

## 1. Introduction

Natural processes constitute optimal solutions in a wide range of complex problems where a larger number of biological cells, organisms or species are involved [8, 35]. Researchers may become inspired by nature to approximate various highly complex, distributed problems, which could involve many dynamic variables and big data [31]. Natural computing [47], which is about using computational techniques inspired by nature, provides new ways to understand, model and analyze environmental complexity, allowing highly accurate approximations of reality and predictions [21].

The penetration of natural computing in the environmental informatics research domain is an open question. This paper investigates the degree to which the research community has been inspired by nature in order to approach some complex environmental problems, examining the potential and perspectives of nature-inspired computing, as well as the types of applications where it has been applied. In the following sections, the aforementioned questions are addressed through a survey of the state of the art work at the field. To our knowledge, this is the first effort that tries to map the nature-inspired research work in the research area of environmental informatics.

## 2. Methodology

In the context of this survey, environmental informatics refers to the science of information applied to environmental science, including urban or rural infrastructures used for the management of physical resources (i.e. water, natural gas and energy). From the environmental science field, this survey mostly focuses on ecology, agriculture, zoology, soil and plant science, and atmospheric science. This review does not cover biology, physics or chemistry, nor urban computing involving humans and their interaction with

---

[1] Department of Computer Science, University of Twente, The Netherlands. Email: a.kamilaris@utwente.nl

the urban landscape. A keyword-based search for related work was performed through the web scientific indexing services Web of Science and Google Scholar. The following query was used:

"Natural Computing" AND ["Environmental Informatics"].

Thirty four (34) papers were initially discovered. In order to increase the number of identified papers, in the place of the keyword "Natural Computing", we tried more specific keywords from the nature-inspired computing research area (see Section 3), such as "Artificial Neural Networks" (ANN), "Swarm Intelligence" (SI), "Deep Learning" (DL) etc. It is noted that some areas of natural computing, such as "Membrane computing" [61], "DNA Computing" [62], "Molecular Computing" [71], "Quantum Computing" [22], or "Physics-Based Algorithms" [80] and "Artificial Immune Systems" [15] did not find (yet) application in environmental research as defined above (to our knowledge), for the scope of this survey. Each paper was checked for relevance based on its abstract, and then placed into one of eight different categories (see Section 3), according to the type of natural computing approach used. Afterwards, each category was analyzed one-by-one, examining all the papers identified at the previous stage for relevance, by reading each paper in detail. Seventy six (76) papers were selected in total, presented in the next section. Wherever possible for reasons of saving space, survey papers were selected to be cited, covering a group of relevant papers together (instead of citing each paper separately).

*We acknowledge the fact that due to limited space, there might have been some related work omitted. We tried to include at least the most popular relevant papers, with number of citations as the main criterion.*

## 3. Applications of Natural Computing in Environmental Informatics

This section divides natural computing into eight different areas, according to the particular nature-inspired technique used. Each of these areas fall under the general umbrella of artificial intelligence. Each sub-section lists relevant work in one of these areas, related to environmental informatics.

### 3.1 Reinforcement Learning

Reinforcement learning (RL) is inspired by behaviorist psychology, concerned with how software agents take actions in an environment so as to maximize some notion of cumulative reward. Agents learn by experience, finding a balance between exploration (of uncharted territory) and exploitation (of current knowledge). Relevant projects in this area include solar panel efficient operation [1] and optimal operation of water reservoirs [9].

### 3.2 Artificial Neural Networks

ANN are computing systems inspired by the biological neural networks that constitute animal brains. Such systems are based on a collection of connected units called neurons and they learn various non-linear functions through (usually labeled) examples.

In the domain under study, they have been used in prediction of wheat yield [73], rainfall [18], existence of macroinvertebrates in freshwaters [20, 59], and contamination by groundwater arsenic [13]. ANN have been also used to model and predict dissolved oxygen in river water [36], hydrologic variables [24], climate change [42], eutrophication in Dutch lakes [68], impact of environmental factors for assessing aquatic ecosystems' quality [58], stress phenotyping in plants [81] and river phytoplankton [29].

Moreover, ANN have been employed to estimate lake water quality using satellite imagery [57], flood frequency at water catchments [79] and deforestation risk [49]. Finally, ANN have been exploited in mapping reefs and sub-tidal rocky habitats [90], precipitation analysis in humid regions to detect drought [88], cough recognition of pigs [11], mapping land cover in tropical coastal areas [50], and for more precise predictions at the wine industry [77].

### 3.3 Deep Learning

DL is about deeper ANN that provide a hierarchical representation of the data by means of various convolutions [39]. This allows larger learning capabilities and thus higher performance and precision. An extensive survey on various applications of DL in agriculture is done in [32], with applications in land cover classification, plants/crops/animals classification, plants/leaves disease detection, fruit counting and weeds identification. Another survey on applications of DL in remote sensing is presented in [7], with applications in animal detection, environment and water analysis, and oil spill detection. DL has also been used for weather monitoring [76], bird song classification [87] and species recognition for wild animal monitoring [12]. Finally, DL has been applied for disaster recognition and monitoring, such as natural disasters [4, 33] and geological disasters [41], using images either from satellites or from drones (i.e. aerial photos).

### 3.4 Swarm Intelligence

The collective behavior of bees, bacteria, glow-worms, fireflies, slime moulds, cockroaches, mosquitoes and other organisms have inspired swarm intelligence researchers to devise new optimization algorithms [60]. These techniques are usually probabilistic and population-based, for combinatorial optimization problems where the search space can be represented by a graph. Artificial agents representing natural organisms (e.g. ants, fireflies) follow links in the graph, exploiting the organisms' behavior of constructing paths between their colony and sources of attraction (e.g. food). In this way, they incrementally discover optimal paths in the graph, which would form the solution. It is noted that this category belongs to the more general area of agent-based modeling [19], which is a rule-based, computational modeling methodology that focuses on interactions among the individual components or agents of some system, generating emerging system behavior.

SI has found applications in estimating health risks of air quality [82], understanding the effect of changing landscape structure on animal species [86], design of water distribution networks [85, 3] and electric power systems [93], understanding outbreaks of the mountain pine beetle [63], prediction of precipitation patterns in South America [75], estimation of land-use and land-cover change [45] with a focus on agricultural land [55], solving the problem of transporting animal manure from livestock to crop farms to be used as fertilizer [30], modeling of mosquito foraging behavior for malaria control [23] and modeling of deforestation [46], as well as classification of natural terrain features [78].

## 3.5 Cellular Automata

Cellular automata (CA) [14] are inspired by complex biological systems and processes, such as snowflakes, seashells and mollusc shells. CA are collections of colored cells of specified shape on a grid that evolve through a number of discrete time steps, according to a set of rules based on the states of neighboring cells. CA have been used in the context of environmental informatics for identification and prediction of landslides [6], understanding of vegetation dynamics [91], modeling the death and reproduction opportunities of trees and grasses [89], modeling land cover change and its impact on watersheds [48], simulating insect outbreaks in forest ecological systems [64] and, finally, for predicting the spatial-temporal evolution of wetlands [92].

## 3.6 Evolutionary Computation and Genetic Algorithms

Evolutionary computation (EC) is a family of algorithms for global optimization inspired by biological evolution, solving problems based on population-based trial and error [17]. An initial set of candidate solutions is generated and updated iteratively. Each new generation is produced by stochastically removing less desired solutions and introducing small random changes. EC has been employed in geo-planning of renewable energy production [43], forecasting of microcystin concentrations in Japan [10], as well as in various Geospatial Information Systems (GIS)-based applications for predicting and modeling climate conditions, rainfall, vegetation type, topography, hydrology, land cover and soils [38].

Genetic algorithms (GA) form a sub-class of EC, generating high-quality solutions to optimization and search problems by relying on bio-inspired operators such as mutation, crossover and selection [5]. They have been used as heuristics in the design and optimization of water distribution networks [84], irrigation water distribution networks [66] and natural gas pipe networks [16]. They have also been harnessed for the modeling of algal blooms in freshwater lakes [67], for predicting potential distributions of invasive fish species in Japan [27] and for optimizing pollution control strategies in watersheds [83].

## 3.7 Fuzzy Logic

Fuzzy logic (FL) uses linguistic variables, defined as fuzzy sets, to approximate human reasoning [51], inspired by how our brains work. It is an approach to computing based on degrees of truth rather than the usual Boolean logic (i.e. true or false). FL has been used for the production of landslide susceptibility maps [65], forecasting of municipal solid waste [34] and seasonal snow-melt runoff [44], for quality evaluation of surface water [26] and water in rivers [56], for prediction of soil erosion in watersheds [52], and for decision support in ecosystem management [2] (i.e. environmental impact assessment of pesticides, impact assessment of fish farming, threatened species classification and integrated environmental management).

Researchers sometimes prefer FL over ANN, when they need a more transparent representation of the system under study [2]. The logical structure of the rules facilitates the understanding and analysis of the model in a semi-qualitative manner, close to the way humans reason about the real world. However, FL has large computing processing and data storage requirements.

## 3.8 Evolutionary Robotics

Although the external structure of robots often does not associate them with nature-inspired research, robots are important platforms for testing many hypotheses in natural computing [47]. Autonomous robots use evolutionary techniques to adapt their behavior to a changing environment; this is the field of evolutionary robotics. Moller et al. [53] described a robot which uses a navigational system based upon that of the desert ant *Cataglyphis fortis* which can navigate effectively under similar conditions to that of the living animal. Robots have been used in precision agriculture, for weed control [40], plant phenotyping [72] or for assuring optimal growing conditions in greenhouses [28], for supervision, monitoring and remote sensing of territories and lands [69], and for applications that require autonomous behavior in difficult or hazardous environments, such as sewers [70], radioactive areas [25] and search and rescue scenarios [54]. Finally, "Swarm Robotics" is an interesting research area, related to the coordination of large numbers of relatively simple robots, inspired by social insects [74]. Robots in this case might form an intelligent swarm, by combining principles from SI and artificial immune systems, which are systems inspired by the human immune system [15].

## 4. Discussion

From the surveyed papers, the nature-inspired techniques mostly used were DL (29 papers[2]), ANN (20 papers), EC-GA (17 papers[3]) and SI (12 papers). DL meets increasing popularity in remote sensing (e.g. for land cover mapping and evaluation of climate change) [7], agriculture [32] and disaster detection [4, 33]. Common applications of environmental research, where natural computing has been applied, include infrastructures for physical resources (i.e. planning of water/energy/gas distribution networks), disaster monitoring and assessment, modeling of climate/weather, as well as analysis/modeling of land use changes, biodiversity analysis and monitoring of wild animals, fish and insects. Table 1 presents the applications of environmental informatics, together with the respective nature-inspired technique(s) used.

| App./Technique | RL | ANN | DL | SI | CA | EC-GA | FL | Robotics |
|---|---|---|---|---|---|---|---|---|
| Renewable energy | X | - | - | - | - | X | - | - |
| Infrastructures for physical resources | X | - | - | X | - | X | - | - |
| Agriculture | - | X | X | X | X | - | X | X |

---

[2] Some of the papers are listed in [7, 32], not directly referenced in this paper.
[3] Some of the papers are listed in [38], not directly referenced in this paper.

| Application | | | | | | | | |
|---|---|---|---|---|---|---|---|---|
| Weather forecasting | - | X | X | X | - | X | X | - |
| Water quality and contamination | - | X | X | - | - | X | X | - |
| Air quality | - | - | - | X | - | - | - | - |
| Climate change | - | X | - | - | - | X | - | - |
| Disaster detection and assessment | - | X | X | - | X | - | X | X |
| Deforestation | - | X | - | X | X | - | - | - |
| Biodiversity | - | X | X | X | - | X | X | - |
| Land cover mapping | - | X | X | X | X | X | - | - |
| Animals, fish and insects monitoring | - | - | X | X | X | X | X | - |
| Prediction of waste produced | - | - | - | - | - | - | X | - |
| Soil erosion | - | - | - | - | - | - | X | - |
| Operations in hazardous environments | - | - | - | - | - | - | - | X |

**Tab. 1.** Applications of environmental informatics and respective nature-inspired techniques used.

In many cases natural computing comes in combination with emerging sensing technologies, and it is used as a second step for classification, regression or modeling. Examples include satellite imagery, GIS and ANN [50, 57], GIS, GPS and ANN [49, 77], satellite imagery and SI [78], GIS and CA [64], and, finally, GIS and EC [38]. Robotics make use of real-world sensors and cameras [70]. A particularly sensitive topic is the one of understanding and modeling climate change, and its possible implications. ANN [42] and EC [38] have been suggested as potential methods to approximate the problem.

Our survey revealed various benefits of applying nature-inspired techniques in environmental informatics. They have low implementation costs and the chances of being entrapped in local minima are very low [37]. In particular, SI is robust and flexible for dynamic problems and environments [60], while ANN and DL approximate well complex non-linear problems. DL captures well variations among image classes without the need of time-consuming and error-prone feature engineering [32]. Further, SI and EC/GA constitute good heuristic solutions in NP-hard computationally complex problems, assisting to find optimal solutions quite fast [17, 5]. Their disadvantages involve that there is no guarantee that the solution found is a global optimum and their performance is very sensitive to parameter setting [37]. For ANN and DL, appropriate large datasets are required, with proper data annotation.

Finally, a prediction of the authors is that more applications of DL are expected, as it appears to be a promising technique that generally outperforms existing approaches when adequate data is available [32]. It could be applied in cases where ANN have been used in the past (see Section 3.2), or in new applications where (mostly) computer vision is involved, such as robotics.

## 5. Conclusions

This paper examines the existing applications of natural computing in environmental informatics, by reviewing related work at the field. The survey conducted has revealed various nature-inspired techniques, such as artificial neural networks, deep learning, swarm intelligence and evolutionary computation, applied in a diverse range of environmental informatics-related problems, such as land cover classification, modeling of natural resources, agricultural processes, prediction of precipitation, effects of climate change, disaster identification etc. Natural computing offers various benefits, such as robustness, flexibility, low implementation costs and good approximation of non-linear complex problems. Its disadvantages include no guarantees for optimal solution, high sensitivity to parameter setting and the need of appropriate datasets in ANN/DL problems. More extensive use of natural computing is expected in the future, especially using the recent, promising technique of deep learning.

## References

[1] David Abel, Emily Reif, Edward C Williams, and Michael L Littman. Toward improving solar panel efficiency using reinforcement learning, 2017.

[2] Veronique Adriaenssens, Bernard De Baets, Peter LM Goethals, and Niels De Pauw. Fuzzy rule-based models for decision support in ecosystem management. Science of the Total Environment, 319(1-3):1–12, 2004.

[3] Kazem Mohammadi Aghdam, Iraj Mirzaee, Nader Pourmahmood, and Mohammad Pourmahmood Aghababa. Design of water distribution networks using accelerated momentum particle swarm optimisation technique. Journal of Experimental & Theoretical Artificial Intelligence, 26(4):459–475, 2014.

[4] Siti Nor Khuzaimah Binti Amit, Soma Shiraishi, Tetsuo Inoshita, and Yoshimitsu Aoki. Analysis of satellite images for disaster detection. In Geoscience and Remote Sensing Symposium (IGARSS), 2016 IEEE International, pages 5189– 5192. IEEE, 2016.

[5] Christine M Anderson-Cook. Practical genetic algorithms, 2005.

[6] Maria Vittoria Avolio, Valeria Lupiano, Paolo Mazzanti, and Salvatore Di Gregorio. A cellular automata model for flow-like landslides with numerical simulations of subaerial and subaqueous cases. EnviroInfo, 1:131–140, 2009.

[7] John E Ball, Derek T Anderson, and Chee Seng Chan. Comprehensive survey of deep learning in remote sensing: theories, tools, and challenges for the community. Journal of Applied Remote Sensing, 11(4):042609, 2017.

[8] S Binitha, S Siva Sathya, et al. A survey of bio inspired optimization algorithms. International Journal of Soft Computing and Engineering, 2(2):137–151, 2012.

[9] A Castelletti, S Galelli, M Restelli, and R Soncini-Sessa. Tree-based reinforcement learning for optimal water reservoir operation. Water Resources Research, 46(9), 2010.

[10] Wai Sum Chan, Friedrich Recknagel, Hongqing Cao, and Ho-Dong Park. Elucidation and short-term forecasting of microcystin concentrations in lake Suwa (japan) by means of artificial neural networks and evolutionary algorithms. Water Research, 41(10):2247–2255, 2007.

[11] Allel Chedad, Dimitrios Moshou, Jean-Marie Aerts, A Van Hirtum, Herman Ramon, and Daniel Berckmans. Apanimal production technology: recognition system for pig cough based on probabilistic neural networks. Journal of agricultural engineering research, 79(4):449–457, 2001.

[12] Guobin Chen, Tony X Han, Zhihai He, Roland Kays, and Tavis Forrester. Deep convolutional neural network based species recognition for wild animal monitoring. In Image Processing (ICIP), 2014 IEEE International Conference on, pages 858–862. IEEE, 2014.

[13] Kyung Hwa Cho, Suthipong Sthiannopkao, Yakov A Pachepsky, Kyoung-Woong Kim, and Joon Ha Kim. Prediction of contamination potential of groundwater arsenic in Cambodia, Laos, and Thailand using artificial neural network. Water research, 45(17):5535–5544, 2011.

[14] B Chopard and M Droz. Cellular automata. Springer, 1998.

[15] Leandro Nunes De Castro and Fernando José Von Zuben. Artificial immune systems: Part i–basic theory and applications. Universidade Estadual de Campinas, Dezembro de, Tech. Rep, 210(1), 1999.

[16] Omar Fayez Mohamed El-Mahdy, Mohamed Ezz Hassan Ahmed, and Sayed Metwalli. Computer aided optimization of natural gas pipe networks using genetic algorithm. Applied Soft Computing, 10(4):1141–1150, 2010.

[17] David B Fogel. Introduction to evolutionary computation. Evolutionary computation, 1:1–3, 2000.

[18] Mark N French, Witold F Krajewski, and Robert R Cuykendall. Rainfall forecasting in space and time using a neural network. Journal of hydrology, 137(1-4):1–31, 1992.

[19] Nigel Gilbert. Agent-based models. Number 153. Sage, 2008.

[20] Peter LM Goethals, Andy P Dedecker, Wim Gabriels, Sovan Lek, and Niels De Pauw. Applications of artificial neural networks predicting macroinvertebrates in freshwaters. Aquatic Ecology, 41(3):491–508, 2007.

[21] David G Green and Nicholas I Klomp. Environmental informatics - a new paradigm for coping with complexity in nature. Complex Systems, 98:36–44, 1998.

[22] Jozef Gruska. Quantum computing, volume 2005. McGraw-Hill London, 1999.

[23] Weidong Gu and Robert J Novak. Agent-based modelling of mosquito foraging behaviour for malaria control. Transactions of the Royal Society of Tropical Medicine and Hygiene, 103(11):1105–1112, 2009.

[24] Kuo-lin Hsu, Hoshin V Gupta, Xiaogang Gao, Soroosh Sorooshian, and Bisher Imam. Self-organizing linear output map (SOLO): An artificial neural network suitable for hydrologic modeling and analysis. Water Resources Research, 38(12), 2002.

[25] Andres Iborra, Juan A Pastor, Barbara Álvarez, Carlos Fernandez, and Jose M Fernandez Merono. Robots in radioactive environments. IEEE Robotics & Automation Magazine, 10(4):12–22, 2003.

[26] Yilmaz Icaga. Fuzzy evaluation of water quality classification. Ecological Indicators, 7(3):710–718, 2007.

[27] Kei'ichiro Iguchi, Keiichi Matsuura, Kristina M McNyset, A Townsend Peterson, Ricardo Scachetti-Pereira, Katherine A Powers, Dave A Vieglais, Edward O Wiley, and Taiga Yodo. Predicting invasions of north American basses in Japan using native range data and a genetic algorithm. Transactions of the American Fisheries Society, 133(4):845–854, 2004.

[28] Simon Janos and Istvan Matijevics. Implementation of potential field method for mobile robot navigation in greenhouse environment with WSN support. In Intelligent Systems and Informatics (SISY), 2010 8th International Symposium on, pages 319–323. IEEE, 2010.

[29] Kwang-Seuk Jeong, Dong-Kyun Kim, and Gea-Jae Joo. River phytoplankton prediction model by artificial neural network: Model performance and selection of input variables to predict time-series phytoplankton proliferations in a regulated river system. Ecological Informatics, 1(3):235–245, 2006.

[30] Andreas Kamilaris, Anton Assumpcio, August Bonmati Blasi, Marta Torrellas, and Francesc X. Prenafeta-Boldu. Estimating the environmental impact of agriculture by means of geospatial and big data analysis: The case of Catalonia. In Proc. of EnviroInfo, Luxembourg, September 2017.

[31] Andreas Kamilaris, Andreas Kartakoullis, and Francesc X. Prenafeta-Boldu. A Review on the Practice of Big Data Analysis in Agriculture. Computers and Electronics in Agriculture International Journal, 143(1):23–37, December 2017.

[32] Andreas Kamilaris and Francesc X. Prenafeta-Boldu. Deep learning in agriculture: A survey. Computers and Electronics in Agriculture, 147:70–90, 2018.

[33] Andreas Kamilaris and Francesc X. Prenafeta-Boldu. Disaster monitoring using unmanned aerial vehicles and deep learning. In Disaster Management for Resilience and Public Safety Workshop, in Proc. of EnviroInfo2017, Luxembourg, September 2017.

[34] Vassilios Karavezyris, Klaus-Peter Timpe, and Ruth Marzi. Application of system dynamics and fuzzy logic to forecasting of municipal solid waste. Mathematics and Computers in simulation, 60(3-5):149–158, 2002.

[35] Lila Kari and Grzegorz Rozenberg. The many facets of natural computing. Communications of the ACM, 51(10):72–83, 2008.

[36] O Kisi, N Akbari, M Sanatipour, A Hashemi, K Teimourzadeh, and J Shiri. Modeling of dissolved oxygen in river water using artificial intelligence techniques. Journal of Environmental Informatics, 22(2), 2013.

[37] Arthur K Kordon. Swarm intelligence: The benefits of swarms. In Applying Computational Intelligence, pages 145–174. Springer, 2010.

[38] Kenneth H Kozak, Catherine H Graham, and John J Wiens. Integrating GIS-based environmental data into evolutionary biology. Trends in ecology & evolution, 23(3):141–148, 2008.

[39] Yann Le Cun, Yoshua Bengio, and Geoffrey Hinton. Deep learning. nature, 521(7553):436, 2015.

[40] Won Suk Lee, DC Slaughter, and DK Giles. Robotic weed control system for tomatoes. Precision Agriculture, 1(1):95–113, 1999.

[41] Ying Liu and Linzhi Wu. Geological disaster recognition on optical remote sensing images using deep learning. Procedia Computer Science, 91:566–575, 2016.

[42] Ze Lin Liu, Chang Hui Peng, Wen Hua Xiang, Da Lun Tian, Xiang Wen Deng, and Mei Fang Zhao. Application of artificial neural networks in global climate change and ecological research: An overview. Chinese science bulletin, 55(34):3853–3863, 2010.

[43] Daniel Luckehe, Oliver Kramer, and Manfred Weisensee. An evolutionary approach to geo-planning of renewable energies. In EnviroInfo, pages 501–508, 2014.

[44] C Mahabir, FE Hicks, and A Robinson Fayek. Application of fuzzy logic to forecast seasonal runoff. Hydrological processes, 17(18):3749–3762, 2003.

[45] Steven M Manson. Agent-based modeling and genetic programming for modeling land change in the southern Yucatan peninsula region of Mexico. Agriculture, ecosystems & environment, 111(1-4):47–62, 2005.

[46] Steven M Manson and Tom Evans. Agent-based modeling of deforestation in southern Yucatan, Mexico, and reforestation in the Midwest United States. Proceedings of the National Academy of Sciences, 104(52):20678–20683, 2007.

[47] Paul Marrow. Nature-inspired computing technology and applications. BT Technology Journal, 18(4):13–23, 2000.

[48] Eric Marshall and Timothy O Randhir. Spatial modeling of land cover change and watershed response using Marconian cellular automata and simulation. Water Resources Research, 44(4), 2008.

[49] Jean-Francois Mas, Henri Puig, Jose Luis Palacio, and Atahualpa Sosa-Lopez. Modelling deforestation using GIS and artificial neural networks. Environmental Modelling & Software, 19(5):461–471, 2004.

[50] JF Mas. Mapping land use/cover in a tropical coastal area using satellite sensor data, GIS and artificial neural networks. Estuarine, Coastal and Shelf Science, 59(2):219–230, 2004.

[51] Jerry M Mendel. Uncertain rule-based fuzzy logic systems: introduction and new directions. Prentice Hall PTR Upper Saddle River, 2001.

[52] B Mitra, HD Scott, J Cetal Dixon, and JM McKimmey. Applications of fuzzy logic to the prediction of soil erosion in a large watershed. Geoderma, 86(3-4):183–209, 1998.

[53] Ralf Moller, Dimitrios Lambrinos, Rolf Pfeifer, Thomas Labhart, and Rudiger Wehner. Modeling ant navigation with an autonomous agent. From animals to animats, 5:185–194, 1998.

[54] Robin R Murphy, Satoshi Tadokoro, Daniele Nardi, Adam Jacoff, Paolo Fiorini, Howie Choset, and Aydan M Erkmen. Search and rescue robotics. In Springer Handbook of Robotics, pages 1151–1173. Springer, 2008.

[55] Dave Murray-Rust, Derek T Robinson, Eleonore Guillem, Eleni Karali, and Mark Rounsevell. An open framework for agent based modelling of agricultural land use change. Environmental modelling & software, 61:19–38, 2014.

[56] William Ocampo-Duque, Nuria Ferre-Huguet, Jose L Domingo, and Marta Schuhmacher. Assessing water quality in rivers with fuzzy inference systems: a case study. Environment International, 32(6):733–742, 2006.

[57] SS Panda, V Garg, and I Chaubey. Artificial neural networks application in lake water quality estimation using satellite imagery. Journal of Environmental Informatics, 4(2):65–74, 2004.

[58] Young-Seuk Park, Tae-Soo Chon, Inn-Sil Kwak, and Sovan Lek. Hierarchical community classification and assessment of aquatic ecosystems using artificial neural networks. Science of the Total Environment, 327(1-3):105–122, 2004.

[59] Young-Seuk Park, Piet FM Verdonschot, Tae-Soo Chon, and Sovan Lek. Patterning and predicting aquatic macroinvertebrate diversities using artificial neural network. Water research, 37(8):1749–1758, 2003.

[60] Rafael S Parpinelli and Heitor S Lopes. New inspirations in swarm intelligence: a survey. International Journal of Bio-Inspired Computation, 3(1):1–16, 2011.

[61] Gheorghe Paun. Membrane computing: an introduction. Springer Science & Business Media, 2012.

[62] Gheorghe Paun, Grzegorz Rozenberg, and Arto Salomaa. DNA computing: new computing paradigms. Springer Science & Business Media, 2005.

[63] Liliana Perez and Suzana Dragicevic. ForestSimMPB: A swarming intelligence and agent-based modeling approach for mountain pine beetle outbreaks. Ecological informatics, 6(1):62–72, 2011.

[64] Liliana Perez and Suzana Dragicevic. Landscape-level simulation of forest insect disturbance: Coupling swarm intelligent agents with GIS-based cellular automata model. Ecological modelling, 231:53–64, 2012.

[65] Biswajeet Pradhan. Use of GIS-based fuzzy logic relations and its cross application to produce landslide susceptibility maps in three test areas in Malaysia. Environmental Earth Sciences, 63(2):329–349, 2011.

[66] Juan Reca and Juan Martınez. Genetic algorithms for the design of looped irrigation water distribution networks. Water resources research, 42(5), 2006.

[67] Friedrich Recknagel, Jason Bobbin, Peter Whigham, and Hugh Wilson. Comparative application of artificial neural networks and genetic algorithms for multivariate time-series modelling of algal blooms in freshwater lakes. Journal of Hydroinformatics, 4(2):125–133, 2002.

[68] Friedrich Recknagel, Anita Talib, and Diederik van der Molen. Phytoplankton community dynamics of two adjacent Dutch lakes in response to seasons and eutrophication control unravelled by non-supervised artificial neural networks. Ecological Informatics, 1(3):277–285, 2006.

[69] Aleksandar Rodic, Dusko Katie, and Gyula Mester. Ambient intelligent robot-sensor networks for environmental surveillance and remote sensing. In Intelligent Systems and Informatics, 2009. SISY'09. 7th International Symposium on, pages 39–44. IEEE, 2009.

[70] Erich Rome, Joachim Hertzberg, Frank Kirchner, Ulrich Licht, and Thomas Christaller. Towards autonomous sewer robots: the MAKRO project. Urban Water, 1(1):57–70, 1999.

[71] Grzegorz Rozenberg, Thomas Bck, and Joost N Kok. Handbook of natural computing. Springer Publishing Company, Incorporated, 2011.

[72] Arno Ruckelshausen, Peter Biber, Michael Dorna, Holger Gremmes, Ralph Klose, Andreas Linz, Florian Rahe, Rainer Resch, Marius Thiel, Dieter Trautz, et al. BoniRob–an autonomous field robot platform for individual plant phenotyping. Precision agriculture, 9(841):1, 2009.

[73] Georg Ruß, Rudolf Kruse, Martin Schneider, and Peter Wagner. Data mining with neural networks for wheat yield prediction. In Industrial Conference on Data Mining, pages 47–56. Springer, 2008.

[74] Erol Sahin. Swarm robotics: From sources of inspiration to domains of application. In International workshop on swarm robotics, pages 10–20. Springer, 2004.

[75] Ariane F dos Santos, Haroldo F de Campos Velho, Eduardo FP Luz, Saulo R Freitas, Georg Grell, and Manoel A Gan. Firefly optimization to determine the precipitation field on south America. Inverse Problems in Science and Engineering, 21(3):451–466, 2013.

[76] Gunjan Sehgal, Bindu Gupta, Kaushal Paneri, Karamjit Singh, Geetika Sharma, and Gautam Shroff. Crop planning using stochastic visual optimization. arXiv preprint arXiv:1710.09077, 2017.

[77] Subana Shanmuganathan, Philip Sallis, Leopoldo Pavesi, and Mary Carmen Jarur Munoz. Computational intelligence and geo-informatics in viticulture. In Modeling & Simulation, 2008. AICMS 08. Second Asia International Conference on, pages 480–485. IEEE, 2008.

[78] Er Sakshi Sharma, Er PoojaNagpal, and Er Harish Kundra. A review on the satellite image classification using swarm intelligence based techniques. International Journal of Advances in Science and Technology (IJAST), 68(68.8):76–481, 2014.

[79] C Shu and TBMJ Ouarda. Flood frequency analysis at ungauged sites using artificial neural networks in canonical correlation analysis physiographic space. Water Resources Research, 43(7), 2007.

[80] Nazmul Siddique and Hojjat Adeli. Nature inspired computing: an overview and some future directions. Cognitive computation, 7(6):706–714, 2015.

[81] Arti Singh, Baskar Ganapathysubramanian, Asheesh Kumar Singh, and Soumik Sarkar. Machine learning for high-throughput stress phenotyping in plants. Trends in plant science, 21(2):110–124, 2016.

[82] Jeetendra Bahadur Singh, Vijay Sena Reddy, Soumya Jana, and Swades De. Assessment of health risk due to PM10 using fuzzy linear membership kriging with particle swarm optimization. In EnviroInfo, pages 887–894, 2013.

[83] P Srivastava, JM Hamlett, PD Robillard, and RL Day. Watershed optimization of best management practices using AnnAGNPS and a genetic algorithm. Water resources research, 38(3), 2002.

[84] Marcin Stachura and Bartlomiej Fajdek. Planning of a water distribution network sensors location for a leakage isolation. In EnviroInfo, pages 715–722, 2014.

[85] CR Suribabu and TR Neelakantan. Particle swarm optimization compared to other heuristic search techniques for pipe sizing. Journal of Environmental Informatics, 8(1), 2006.

[86] Chris J Topping, Tine S Hansen, Thomas S Jensen, Jane U Jepsen, Frank Nikolajsen, and Peter Odderskær. Almass, an agent-based model for animals in temperate European landscapes. Ecological Modelling, 167(1-2):65–82, 2003.

[87] Balint Pal Toth and Balint Czeba. Convolutional neural networks for large-scale bird song classification in noisy environment. In CLEF (Working Notes), pages 560–568, 2016.

[88] Mohammad Valipour. Optimization of neural networks for precipitation analysis in a humid region to detect drought and wet year alarms. Meteorological Applications, 23(1):91–100, 2016.

[89] Mark T Van Wijk and Ignacio Rodriguez-Iturbe. Tree-grass competition in space and time: Insights from a simple cellular automata model based on ecohydrological dynamics. Water Resources Research, 38(9), 2002.

[90] Michael J Watts, Yuxiao Li, Bayden D Russell, Camille Mellin, Sean D Connell, and Damien A Fordham. A novel method for mapping reefs and subtidal rocky habitats using artificial neural networks. Ecological Modelling, 222(15):2606–2614, 2011.

[91] Fei Ye, Qiuwen Chen, and Ruonan Li. Modelling the riparian vegetation evolution due to flow regulation of Lijiang river by unstructured cellular automata. Ecological informatics, 5(2):108–114, 2010.

[92] Huan Yu, Zheng-wei He, Shu-qing Zhang, and Xin Pan. Spatial-temporal evolution simulation of wetland landscape in Sanjiang plain using cellular automaton [j]. Geography and Geo-Information Science, 4:022, 2010.

[93] Xiao-hui Yuan, Cheng Wang, Yong-chuan Zhang, and Yan-bin Yuan. A survey on application of particle swarm optimization to electric power systems [j]. Power System Technology, 19:003, 2004.